\documentclass{ieeeaccess}
\usepackage{cite}
\usepackage{amsmath,amssymb,amsfonts}
\usepackage{algorithmic}
\usepackage{graphicx}
\usepackage{textcomp}
\usepackage{color,soul}

\usepackage{comment}

\usepackage{multirow}
\usepackage{comment}

\def\BibTeX{{\rm B\kern-.05em{\sc i\kern-.025em b}\kern-.08em
    T\kern-.1667em\lower.7ex\hbox{E}\kern-.125emX}}

\begin{document}
\history{Date of publication xxxx 00, 0000, date of current version xxxx 00, 0000.}
\doi{10.1109/ACCESS.2017.DOI}

\title{Fast and Accurate 3D Hand Pose Estimation via Recurrent Neural Network for Capturing Hand Articulations}
\author{\uppercase{Cheol-Hwan Yoo}\authorrefmark{1}, \uppercase{Seo-Won Ji}\authorrefmark{1}, \uppercase{Yong-Goo Shin}\authorrefmark{1}, \uppercase{Seung-Wook Kim}\authorrefmark{1}, and \uppercase{Sung-Jea Ko}\authorrefmark{1}, \IEEEmembership{Fellow, IEEE}}
\address[1]{School of Electrical Engineering, Korea University,
Anam-dong 5(o)-ga, Seongbuk-gu, Seoul, 136-713, Rep. of Korea (e-mail:
chyoo@dali.korea.ac.kr; swji@dali.korea.ac.kr; ygshin@dali.korea.ac.kr; swkim@dali.korea.ac.kr; sjko@korea.ac.kr)}

\markboth
{C.-H. Yoo \headeretal: Fast and Accurate 3D Hand Pose Estimation via Recurrent Neural Network for Capturing Hand Articulations}
{C.-H. Yoo \headeretal: Fast and Accurate 3D Hand Pose Estimation via Recurrent Neural Network for Capturing Hand Articulations}

\corresp{Corresponding author: Sung-Jea Ko (e-mail: sjko@korea.ac.kr).}

\begin{abstract}
3D hand pose estimation from a single depth image plays an important role in computer vision and human-computer interaction. Although recent hand pose estimation methods using convolution neural network~(CNN) have shown notable improvements in accuracy, most of them have a limitation that they rely on a complex network structure without fully exploiting the articulated structure of the hand. A hand, which is an articulated object, is composed of six local parts: the palm and five independent fingers. Each finger consists of sequential-joints that provide constrained motion, referred to as a kinematic chain.  In this paper, we propose a hierarchically-structured convolutional recurrent neural network~(HCRNN) with six branches that estimate the 3D position of the palm and five fingers independently. The palm position is predicted via fully-connected layers. Each sequential-joint, i.e. finger position, is obtained using a recurrent neural network~(RNN) to capture the spatial dependencies between adjacent joints. Then the output features of the palm and finger branches are concatenated to estimate the global hand position. HCRNN directly takes the depth map as an input without a time-consuming data conversion, such as 3D voxels and point clouds. Experimental results on public datasets demonstrate that the proposed HCRNN not only outperforms most 2D CNN-based methods using the depth image as their inputs but also achieves competitive results with state-of-the-art 3D CNN-based methods with a highly efficient running speed of 285 fps on a single GPU.
\end{abstract}


\titlepgskip=-15pt

\maketitle

\section{Introduction}
\label{sec:introduction}

\Figure[t!](topskip=0pt, botskip=0pt, midskip=0pt)[width=0.99\linewidth]{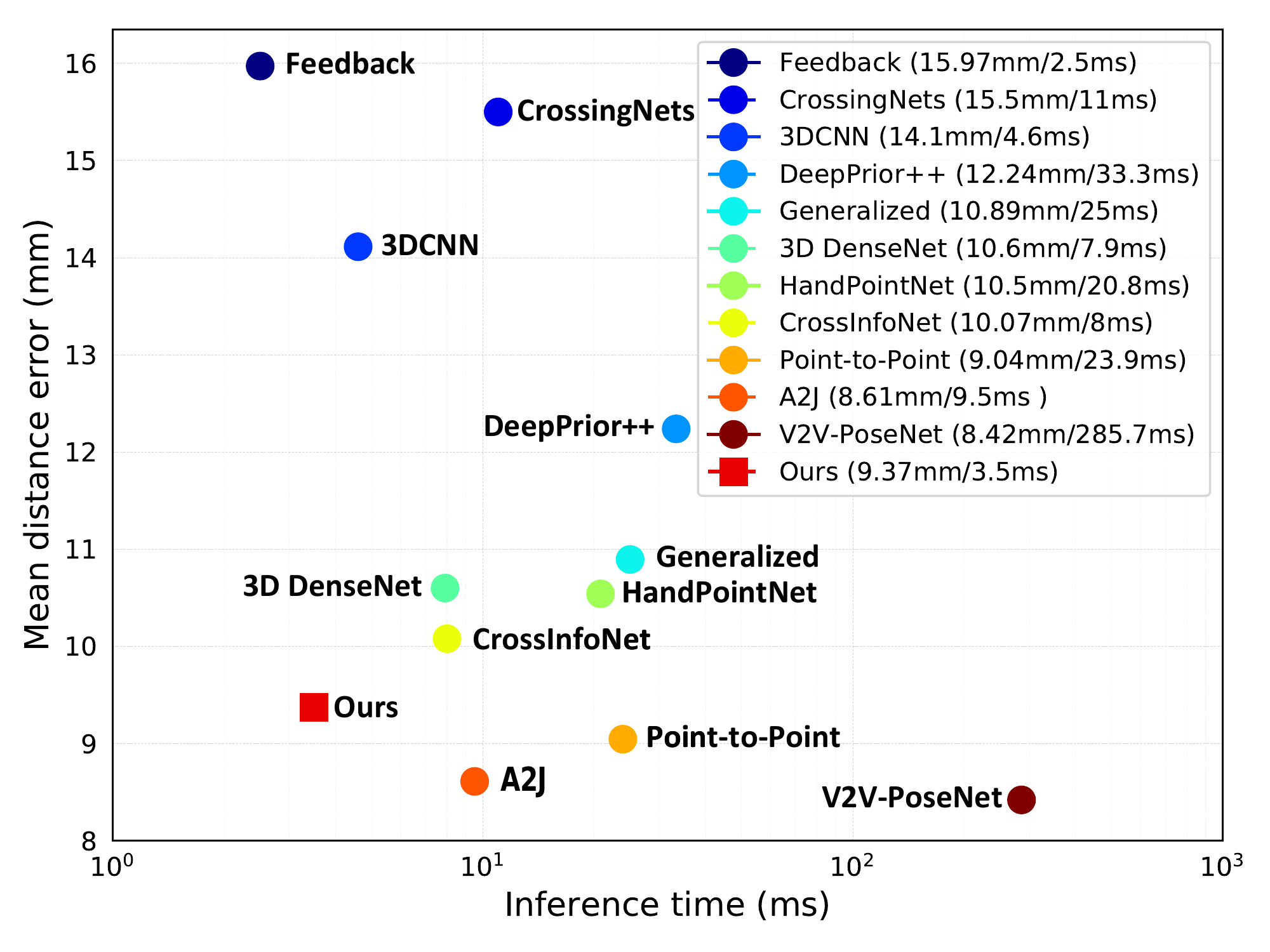}
{Performance/inference speed trade-off on NYU dataset. The proposed HCRNN has advantages both in estimation accuracy and fast inference speed.\label{fig:fig_error_vs_time}}

\Figure[t!](topskip=0pt, botskip=0pt, midskip=0pt)[width=1.0\linewidth]{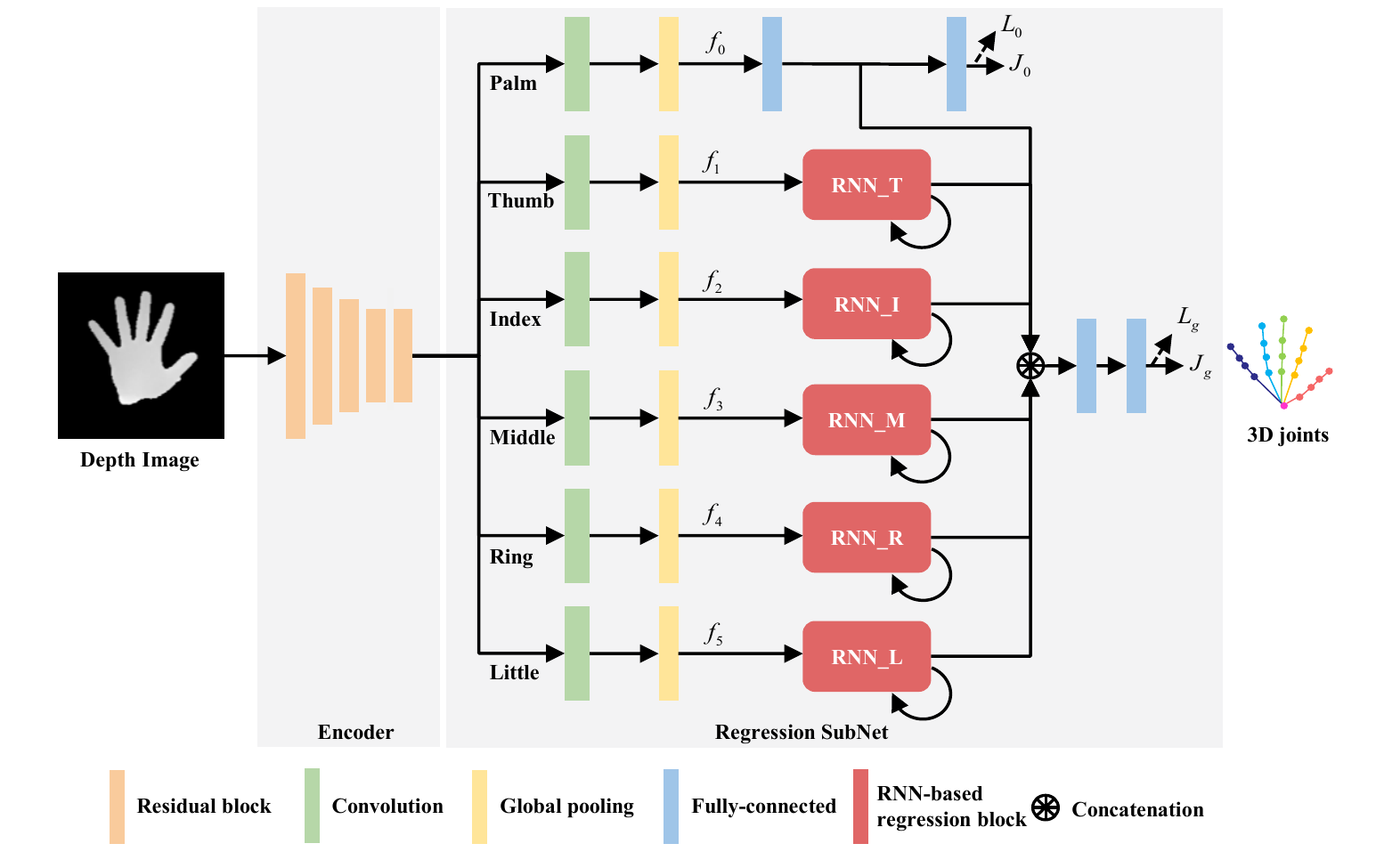}
{The overall architecture of the proposed method for 3D hand pose estimation. It is designed with a hierarchical structure, where each separate branch estimates the 3D positions of the palm and five fingers. In the finger branches, the RNN-based regression block is added to capture the spatial dependencies between the sequential-joints of the finger.\label{fig:framework}}

\PARstart{A}{ccurate} 3D hand pose estimation has received considerable attention regarding a wide range of applications, such as virtual/augmented reality and human-computer interaction~\cite{erol2007vision}. As commercial depth cameras have been released and become more common, depth-based hand pose estimation methods have attracted significant research interest in recent decades~\cite{yuan2018depth}. Nevertheless, it is still a challenging problem to accurately estimate 3D hand pose, because of the low quality of depth images, large variations in hand orientations, high joint flexibility, and severe self-occlusion.

Recently, most 3D hand pose estimation methods have been based on convolutional neural networks~(CNNs) with a single depth image. In these conventional CNN-based methods, there are two major approaches to improve estimation accuracy. 
The first involves 3D data representation of 2D depth images. To utilize 3D spatial information, Ge~\textit{et~al.} and Moon~\textit{et~al.} converted a depth image into a volumetric representation, such as 3D voxels, and then applied a 3D CNN for 3D hand pose estimation~\cite{ge20173d, ge2018real, moon2018v2v}. In addition, 3D representation of inputs based on a 3D point cloud has been proposed~\cite{chen2018shpr, ge2018hand, ge2018point}. Although these methods are effective for capturing the geometric properties of depth images~\cite{ge20173d, moon2018v2v}, they suffer from heavy parameters and complex data-conversion processes, resulting in high time complexity. For efficient training and testing, 2D CNN-based methods that attempt to extract more information from 2D inputs are still being widely researched. In this study, we adopt a 2D depth image itself as input without a further data representation process to utilize the efficiency of 2D CNN.

The second approach involves effective network architecture designs that utilize the structural properties of hands~\cite{du2019crossinfonet, madadi2017end, sun2015cascaded, zhou2018hbe}. For a network model, hierarchical networks that divide the global hand pose estimation problem into sub-tasks have been proposed, where each one focuses on a specific finger or hand region. Sun~\textit{et~al.}~\cite{sun2015cascaded} divided global hand pose estimation into local estimations of palm and finger pose and then updated the finger location according to the palm pose in a cascade manner. Madadi~\textit{et~al.}~\cite{madadi2017end} designed a hierarchically tree-like, structured CNN using five branches to model the five fingers and an additional branch to model the palm orientation. Zhou~\textit{et~al.}~\cite{zhou2018hbe} designed a three-branch network, where the three branches correspond to the thumb, index finger, and the three other fingers, according to the differences in the functional importance of different fingers. More recently, Du~\textit{et~al.}~\cite{du2019crossinfonet} proposed a two-branch cross-connection network that hierarchically regresses palm and finger poses through information-sharing in a multi-task setup. These studies have demonstrated that handling different parts of the hand via a multi-branch CNN can improve the accuracy of 3D hand pose estimation. However, these methods estimate all joints of the finger directly without explicitly considering finger kinematics. For a finger, the movements of different joints in the finger are dependent on each other and can be represented as a kinematic chain. To better capture spatial dependencies between adjacent joints, in our work, we adopt a recurrent neural network~(RNN), which is mainly used to handle the sequential features of the joints in a finger. 

Fig.~\ref{fig:fig_error_vs_time} depicts the inference time~(ms) versus mean distance error~(mm) graph of some state-of-the-art 3D hand pose estimation methods on NYU dataset.
Note that, in terms of estimation accuracy, the proposed method is ranked in the 4th place behind V2V-PoseNet~\cite{moon2018v2v}, A2J~\cite{xiong2019a2j}, and Point-to-Point~\cite{ge2018point} which have much higher computational complexity than the proposed method.
The proposed HCRNN operates approximately 80, seven, and three times faster than V2V-PoseNet~\cite{moon2018v2v}, Point-to-Point~\cite{ge2018point}, and A2J~\cite{xiong2019a2j}, respectively. 
On the other hand, in terms of inference speed, the proposed HCRNN is in the 2nd place behind Feedback~\cite{oberweger2015training}, while HCRNN improves the estimation accuracy of Feedback about 40\%.
In summary, the HCRNN achieves both goals, effectiveness and efficiency, in 3D hand pose estimation. 

Fig.~\ref{fig:framework} illustrates the proposed hierarchical convolutional RNN~(HCRNN), which takes the 3D geometry of a hand into account for 3D hand pose estimation. The six separate branches are based on the observation that the hand is composed of six local parts (i.e. the palm and five fingers) with different amounts of variations due to the articulated structure of the hand.
Inspired by the recent study in which sequential features of a sequence-like object are obtained in a single image~\cite{shi2016end}, we first extract the sequential features of joints by applying the combination of convolutional and fully-connected~(FC) layers to the feature from the encoder network.
Then, we propose to make use of an RNN taking these joint features as inputs to capture the sequential information of a finger and to extract the interdependent information of the finger joints along the kinematic chain.

Our contributions can be summarized as follows:
\begin{enumerate}
\item We propose the HCRNN architecture that decomposes global hand estimation into sub-tasks of estimating the local parts of the hand. Based on the understanding that the palm and finger exhibit different flexibilities and degrees of freedom, the separate branches estimate the 3D positions of the palm and five fingers. We apply the RNN to utilize spatial information between the sequential-joints of the finger.
\item We design a relatively efficient network that not only achieves promising performance with the mean errors of 6.5, 9.4, and 7.7 on the ICVL~\cite{tang2014latent}, NYU~\cite{tompson2014real}, and MSRA~\cite{sun2015cascaded} datasets, respectively, but also runs fast, at over 285 fps, on a single GPU. The speed versus accuracy trade-off graph can be seen in Fig.~\ref{fig:fig_error_vs_time}.
\end{enumerate}

\section{Recurrent Neural Networks and Its Variants}
RNNs learn a hidden representation for each time step of sequential data by considering both the current and previous information. Thanks to their ability to memorize and abstract the sequential information over time, RNN has achieved great success in sequential data modelings such as natural language processing~(NLP) and speech recognition. Formally, the hidden state and the output feature at the current time step, $t$, can be respectively obtained by
\begin{align}
h_{t} &= \mathrm{tanh}(W_{h} x_{t} + U_h h_{t-1} + b_h), \label{eq:1}\\
y_{t} &= W_{y} h_{t} + b_y,\label{eq:2}
\end{align}
where $W_h$, $U_h$, and $b_h$ are the parameters for the hidden state and $W_y$ and $b_y$ are the parameters for the current output. This recurrent structure allows the RNN to convey the information in the past time steps to the current prediction process. However, as the time gap between information grows, the basic RNN cannot preserve temporal memories and faces the problem of long-term dependencies due to the vanishing gradient~\cite{hochreiter2001gradient}. 

To tackle the aforementioned problem, long short-term memory~(LSTM)~\cite{hochreiter1997long} has recently been proposed, which replaces the nonlinear units of the basic RNN. Among the numerous variants of LSTM, the gated recurrent unit~(GRU)~\cite{cho2014learning} is one of the most popular modules owing to its ability to reduce the complexity of LSTM.
There are two key components in GRU, which are referred to as the update gate and reset gate. The update gate, $z_{t}$, controls the balance between the previous and current feature information, while the reset gate, $r_{t}$, is used to modulate the states of the previous hidden feature. At the time step $t$, the current gate output $y_{t}$ is computed as follows:
\begin{align}
r_{t} &= \sigma(W_{r}x_{t}+U_{r}h_{t-1}+b_{r}), \label{eq:3}\\
z_{t} &= \sigma(W_{z}x_{t}+U_{z}h_{t-1}+b_{z}), \label{eq:4}\\
\widetilde{h}_{t} &= \mathrm{tanh}(W_h x_{t}+r_{t}\odot U_h h_{t-1}+b_h), \label{eq:5}\\
h_{t} &= z_{t}\odot h_{t-1}+(1-z_{t})\odot\widetilde{h_{t}}, \label{eq:6}\\
y_{t} &= W_{y} h_{t} + b_y, \label{eq:7}
\end{align}
where $W$, $U$, and $b$ terms with a specific subscript represent the parameters for each layer, respectively, $\sigma (\cdot)$ is a sigmoid function, $\widetilde{h_{t}}$ is the current memory content, $h_t$ is a current hidden state, and $\odot$ represents element-wise multiplication. 
In this work, we adopt a GRU as a basic RNN module.

\section{Proposed Method}
\subsection{Overall Network Architecture}
Fig.~\ref{fig:framework} illustrates the overall architecture of the proposed 3D hand pose estimation methods. The proposed network mainly consists of two parts: an encoder that transforms an input depth image onto the abstracted feature space; a joint regression sub-networks~(SubNets) that are composed of six branches corresponding to five fingers and a palm.
The input depth image is firstly fed into an encoder for low-level feature extraction. Then, the regression SubNets take the obtained feature map from the encoder as an input and predict the 3D pose of a palm and fingers.

\begin{table}[t]
\caption{Detailed architecture of the encoder network for initial feature extraction.}
\vspace{-0.3cm}
	\label{tab1}
	\normalsize
	\begin{center}
		\begin{tabular}{l c c c}
			\hline
			Layers & Kernel size & \# channels & Output size \\
			\hline
			Residual block & $3\times3$ & 64 & 96 \\
			Average pooling & $2\times2$ & 64 & 48 \\
			Residual block & $3\times3$ & 64 & 48 \\
			Average pooling & $2\times2$ & 64 & 24 \\
			Residual block & $3\times3$ & 128 & 24 \\
			Average pooling & $2\times2$ & 128 & 12 \\
			Residual block & $3\times3$ & 256 & 12 \\
			Residual block & $3\times3$ & 256 & 12 \\
			\hline
		\end{tabular}
	\end{center}
\end{table}

\subsection{Encoder Network}
The encoder of the proposed 3D hand pose estimation method is based on ResNet with full pre-activation~\cite{he2016identity} as described in Table~\ref{tab1}. The encoder has five residual blocks, each of which consists of two $3 \times 3$ convolutional layers. For the skip connection in the residual block, we use a $1 \times 1$ convolutional shortcut. Average-pooling layers for down-sampling are appended after each residual block except for the last two blocks. We design the encoder to be shallow because an input depth map has a plainer texture as compared with the inputs used in classification or segmentation tasks. Unless otherwise noted, the encoder takes the input with a spatial size of $96 \times 96$, and thus the output feature map has a spatial size of $12 \times 12$ with 256 channels. 

\subsection{Regression SubNet}
As the different parts of the hands have different amounts of variation and degrees of freedom~(DoF) due to the articulated structure of the hand~\cite{sun2015cascaded}, it is inefficient directly regressing all parts together from the encoded feature. Among hand joints, the palm is much more stable than fingers and mainly affects global hand positions. The five fingers are largely independent and have more flexibility than palm~\cite{sun2015cascaded}. By decomposing the global hand pose estimation problem into sub-tasks of palm and finger pose estimation, the optimization of the network parameters can be simplified by limiting the search space~\cite{ferrari2008progressive, sinha2016deephand}. Based on the aforementioned properties of hands, we design our joint regression SubNet as hierarchically structured networks with six separate branches for the estimation of the palm and five fingers: thumb, index, middle, ring, and little finger.

\subsubsection{Sequential modeling of finger joints}
Although recent state-of-the-art 3D hand pose estimation methods mostly adopt discriminative learning-based methods as deep learning technology advances, model-based methods still have their own advantages~\cite{wohlke2018model, wu2001capturing, zhou2016model}.
As considering the joint connection over the finger, the movements of different joints of a finger are highly related to each other. With a finger root, i.e. metacarpophalangeal~(MCP), as the base position, each finger is composed of sequential-joints, which can be represented as a kinematic chain.
Using the kinematic structure of a hand, the model-based methods can constrain the solution space of the hand joint positions.
To take the advantages of both the model- and discriminative-learning-based methods, we estimate a $n$th joint of the $k$th finger, $J_k^{(n)}$, by using the feature sequence for the previous links as follows:
\begin{equation}
\label{eq:8}
J_k^{(n)} = \mathcal{F}_k^{(n)}\left(f_k^{(n)}, f_{k}^{(n-1)}, \cdots, f_{k}^{(1)}\right),
\end{equation}
where $\mathcal{F}_{k}^{(n)}\left(\cdot\right)$ is a mapping function from the feature sequence onto the 3D coordinate and $f_{k}^{(i)}$, $i=1,\cdots,n$, is the abastracted feature of the $k$th finger's joint. Note that $f_k^{(1)}$ represents the feature of the MCP joint.
Reflecting the kinematic structure of the finger, we design a regression sub-network~(SubNet) of (\ref{eq:8}) by using a recurrent model in which the hidden layer containing the information of previous links controls the current estimation as follows:
\begin{align}
    h_k^{(n)} &= \mathcal{G}\left(h_{k}^{(n-1)}, f_k^{(n)}; \Theta_k\right), \label{eq:9}\\
    J_k^{(n)} &= W_k h_k^{(n)} + b_k, \label{eq:10}
\end{align}
where $h_k^{(n)}$ and $h_{k}^{(n-1)}$ are the current and previous hidden states, respectively, $\mathcal{G}(\cdot)$ is a GRU, $\Theta_k$ is the parameter of the GRU, and $W_k$ and $b_k$ are the parameter for the linear regression.

\Figure[t!](topskip=0pt, botskip=0pt, midskip=0pt)[width=0.99\linewidth]{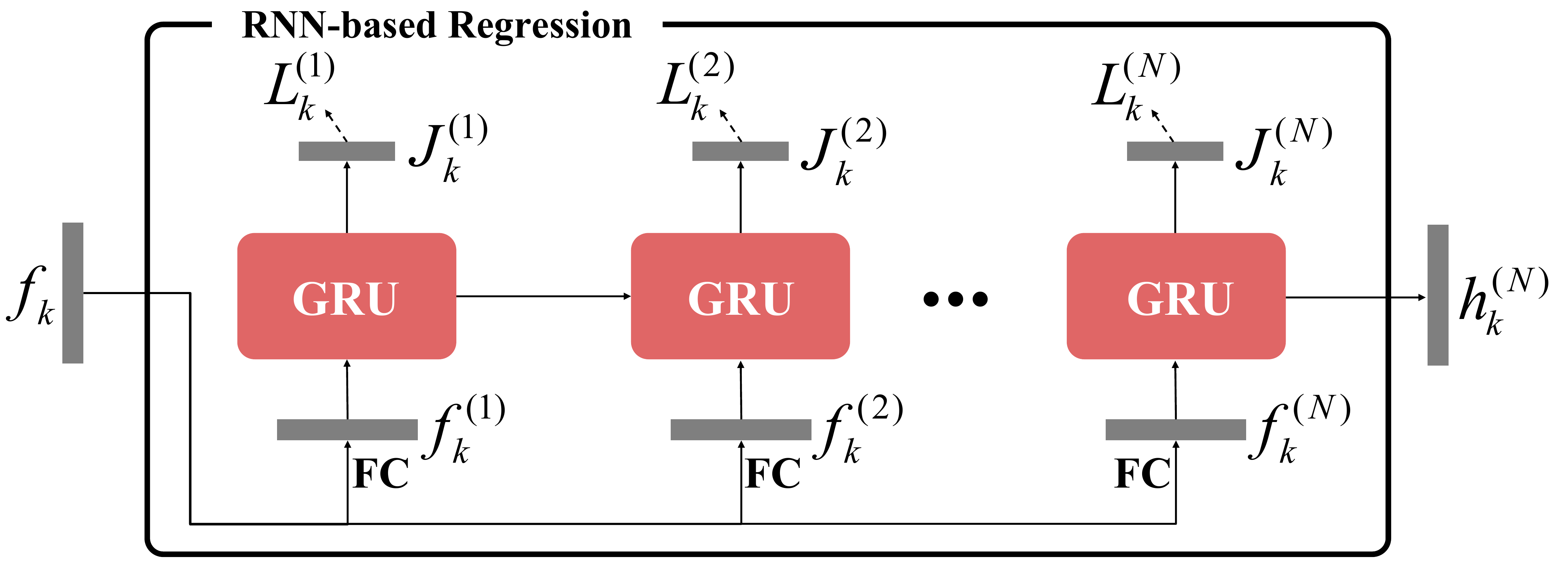}
{Architecture of the RNN-based regression block, which takes a feature sequence of a finger as input. It conveys the feature information of the spatially adjacent joints, which are utilized for estimating the position of sequential finger joints.\label{fig:RNN}}

\subsubsection{Finger branch}
Fig.~\ref{fig:RNN} depicts the RNN-based regression block in the finger branch. For a finger, each 3D joint position is recurrently estimated by using its input encoded feature and the previous joint information in the latent space.
Inspired by the recent study where sequential features of a sequence-like object are obtained in a single image~\cite{shi2016end}, we first extract the sequential features of joints from the output of the encoder network.
To this end, we first apply a $3 \times 3$ convolution with 256 channels followed by a global average pooling to the feature map from the encoder for each branch in order to extract the hand-part-specific information. For example, let $f_k$ be such a feature vector of $k$th finger branch. Taking $f_k$ as an input, $N$ different FC layers are employed to obtain the feature vectors of each joint, $f_k^{(1)},\cdots,f_k^{(N)}$, where $N$ is the number of joints in the finger, as shown in Fig.~\ref{fig:RNN}. 
To estimate the first joint of the finger $J_k^{(1)}$, i.e. the MCP joint, $f_{k}^{(1)}$ and a zero vector as an initial hidden state is fed into the GRU as in (\ref{eq:9}) and (\ref{eq:10}). Then, the hidden state in the GRU conveys the sequential information to the end joint along the kinematic chain. The 3D coordinates of adjacent finger joints $J_k^{(n)}$ for $n=1,\cdots,N$ are sequentially estimated by utilizing the previous hidden state. 

\subsubsection{Palm branch}
Like the finger branches, the hand-part-specific feature $f_0$ for the palm branch is first extracted.
As mentioned at the beginning of this section, the palm is inflexible and more stable than fingers.
Thus, the palm joint $J_{0}$ is directly estimated by applying a series of FC layers to the palm feature $f_{0}$.

\subsubsection{Ensemble strategy}
To incorporate the inter-finger correlation, instead of directly concatenating the joint predictions of each branch, we adopt the feature ensemble strategy as in~\cite{zhou2018hbe, du2019crossinfonet, madadi2017end}.
The features from the last FC layers of the palm branch and the hidden states $h_k^{(N)}$ for $k=1,\cdots,5$ containing the previous joints information from the finger branches are concatenated and followed by 1,024 dimension FC layers to estimate the global 3D hand position $J_g$. 

\subsection{Loss Functions}
To train the proposed network, we adopt the smooth $\ell_1$ loss~\cite{girshick2015fast} defined as 
\begin{equation}
  \mathrm{smooth}_{\ell_{1}}(x) = \begin{cases}
        0.5x^{2}, & \text{if} \ \left | x \right | < 0.01, \\
        0.01(\left | x-0.005\right|), & \text{otherwise},
    \end{cases}
\end{equation} 
In the joint regression SubNet, we define the global joint regression loss $L_{g}$ along with the local joint regression loss $L_{l}$ as:
\begin{equation}
\label{eq:global regression_loss}
L_{g}=\sum_{i=1}^{T}\left|\widetilde{J}_{g}^{(i)}-J_{g}^{(i)}\right|_{smooth},
\end{equation}
\begin{equation}
\label{eq:local regression_loss}
L_l=L_0+\sum_{k=1}^{5}\sum_{n=1}^{N}\left|\widetilde{J}_k^{{(n)}}-J_k^{(n)}\right|_{smooth},
\end{equation}
with 
\begin{equation}
  \left|y\right|_{smooth} = \sum_{j}\mathrm{smooth}_{\ell_{1}}(y_j),
\end{equation} 
where $T$ and $N$ are the number of whole joints and the number of joints in the finger, respectively, $y_j$ is the $j$th component of $y$, and $L_0$ is a palm joint regression loss. $\widetilde{J}_g^{{(i)}}$ and $J_g^{(i)}$ represent the  ground  truth  and  estimated  3D coordinates of $i$th hand joint, $\widetilde{J}_k^{{(n)}}$ and $J_k^{(n)}$ denote the  ground  truth  and  estimated  3D coordinates of $n$th joint from the $k$th finger branch.
The total loss function is defined as:
\begin{equation}
\label{eq:regression_loss} 
L=L_g+\lambda L_l,
\end{equation}
where $\lambda$ is a weight factor to balance the two losses. In our experiment, $\lambda$ is set to 1. 
\Figure[t!](topskip=0pt, botskip=0pt, midskip=0pt)[width=0.9\linewidth]{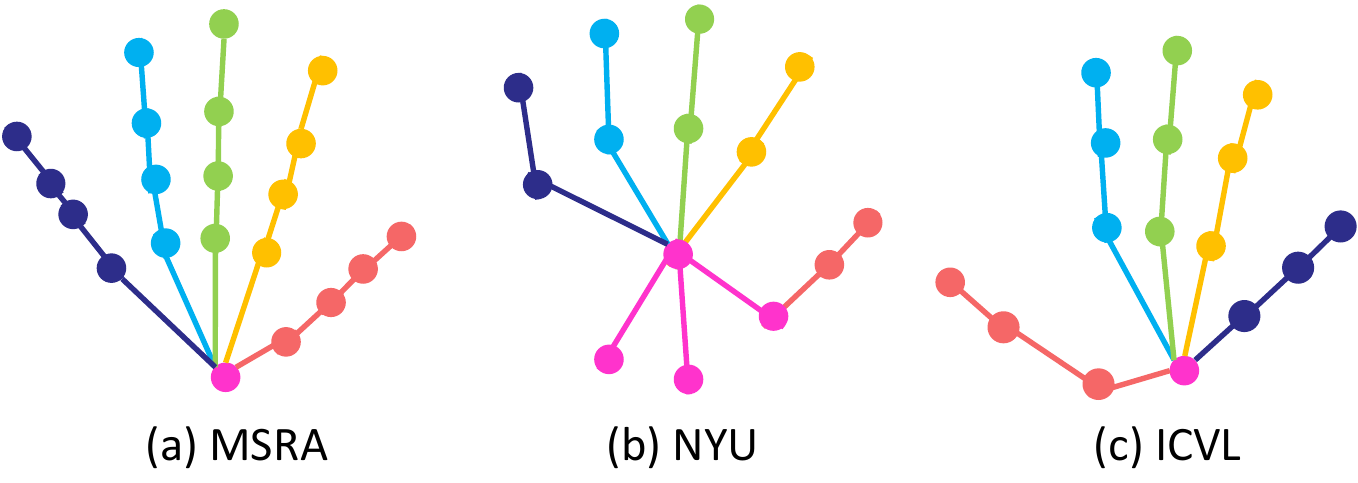}
{Subset of joints on the MSRA, NYU, and ICVL datasets. Violet color indicates the palm joints subset, and other colors indicate the finger joints subset.\label{fig:subset}}

\subsection{Implementation Details}
\label{subsec:Implement}

As mentioned in the previous subsection, the input feature vectors of each branch $f_k$ are obtained by $3 \times 3$ convolutional layers with 256 channels followed by global average pooling layers. The number of output dimensions of the FC layers in the palm branch and the RNN-based regression block is set to 256 except those of the last FC layer for joint regression and the ensemble layer with 1,024 dimensions. We add a batch normalization~(BN) layer after each convolution layer to simplify the learning procedure and improve the generalization ability. The rectified linear unit~(ReLU) is employed as an activation function after the convolutional and FC layers except for the last FC layer, which performs final joint regression. 
Following the strategies in~\cite{chen2019pose, guo2017region, oberweger2017deepprior++, oberweger2015training}, we first extract a fixed-size cube from the depth image around the hand. A hand region is cropped from this bounding area and resized to a fixed size of $96 \times 96$. The depth values within the cropped region are normalized to [-1, 1]. The points for which the depth is outside the range of the cube are assigned a depth of 1. During training, we apply the following online data augmentation tricks: Random rotation in the range [-180$^{\circ}$, 180$^{\circ}$]; random translation of [-10, 10] pixels; random scaling of [0.9, 1.1]. We use the Adam optimizer~\cite{kingma2014adam} with an initial learning rate of 1e-3, a batch size of 32, and a weight decay of 1e-5. The learning rate is multiplied by a factor of 0.96 at every 2k iterations. The entire network is trained for 120 epochs in an end-to-end manner. Our model is implemented by Tensorflow~\cite{abadi2016tensorflow}, and a single NVIDIA Titan X GPU~(Pascal architecture) is used for training and testing. To design the branch details for different hand pose datasets, we define the joint subsets as shown in Fig.~\ref{fig:subset}.

\Figure[t!](topskip=0pt, botskip=0pt, midskip=0pt)[width=0.99\linewidth]{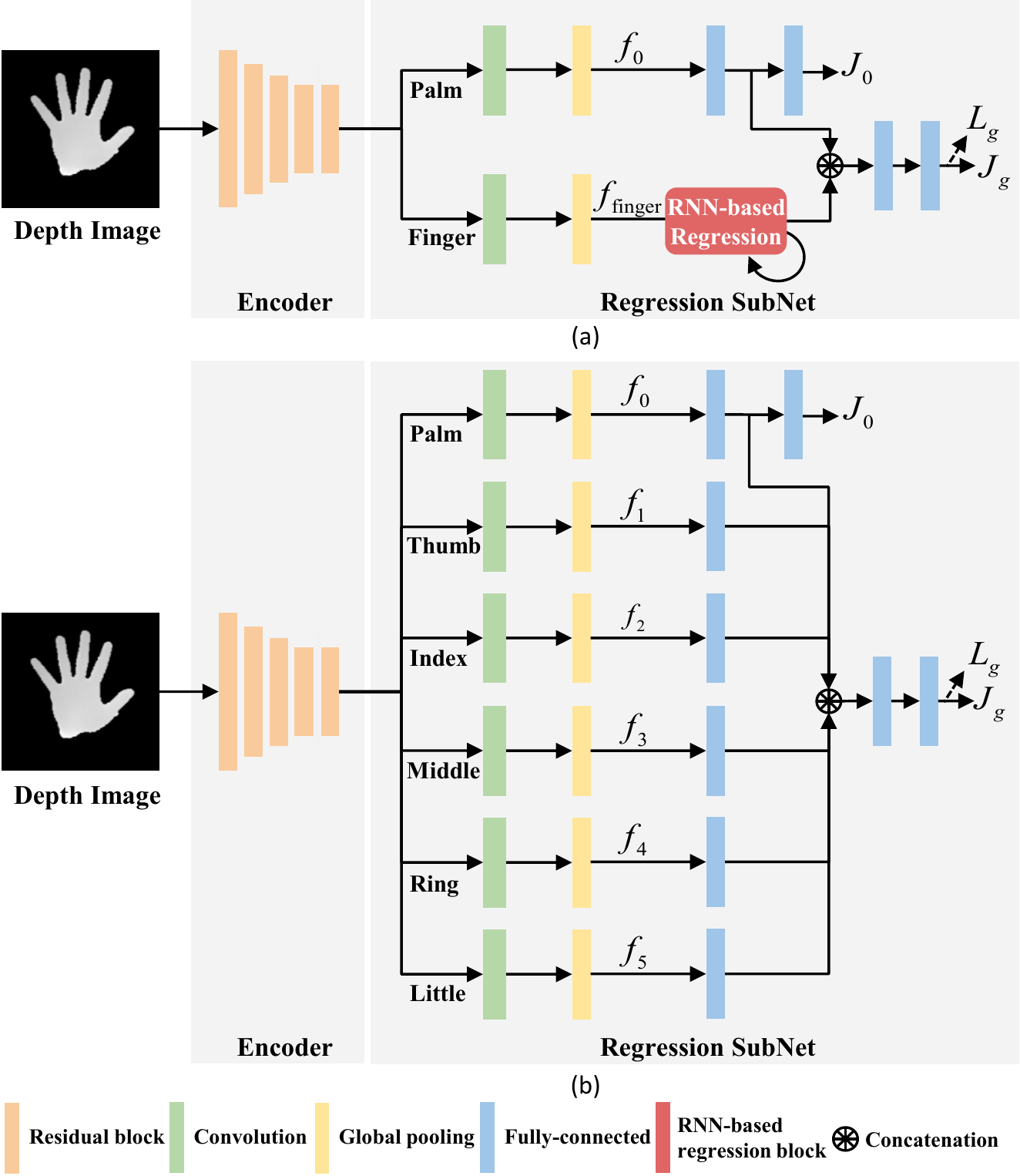}
{Baseline architectures for self-comparisons. a) Two-branch baseline network with one branch for palm and the other branch for finger joint regression. b) Baseline network where the RNN-based regression block is replaced with fully connected layers.\label{fig:self}}

\Figure[t!](topskip=0pt, botskip=0pt, midskip=0pt)[width=0.85\linewidth]{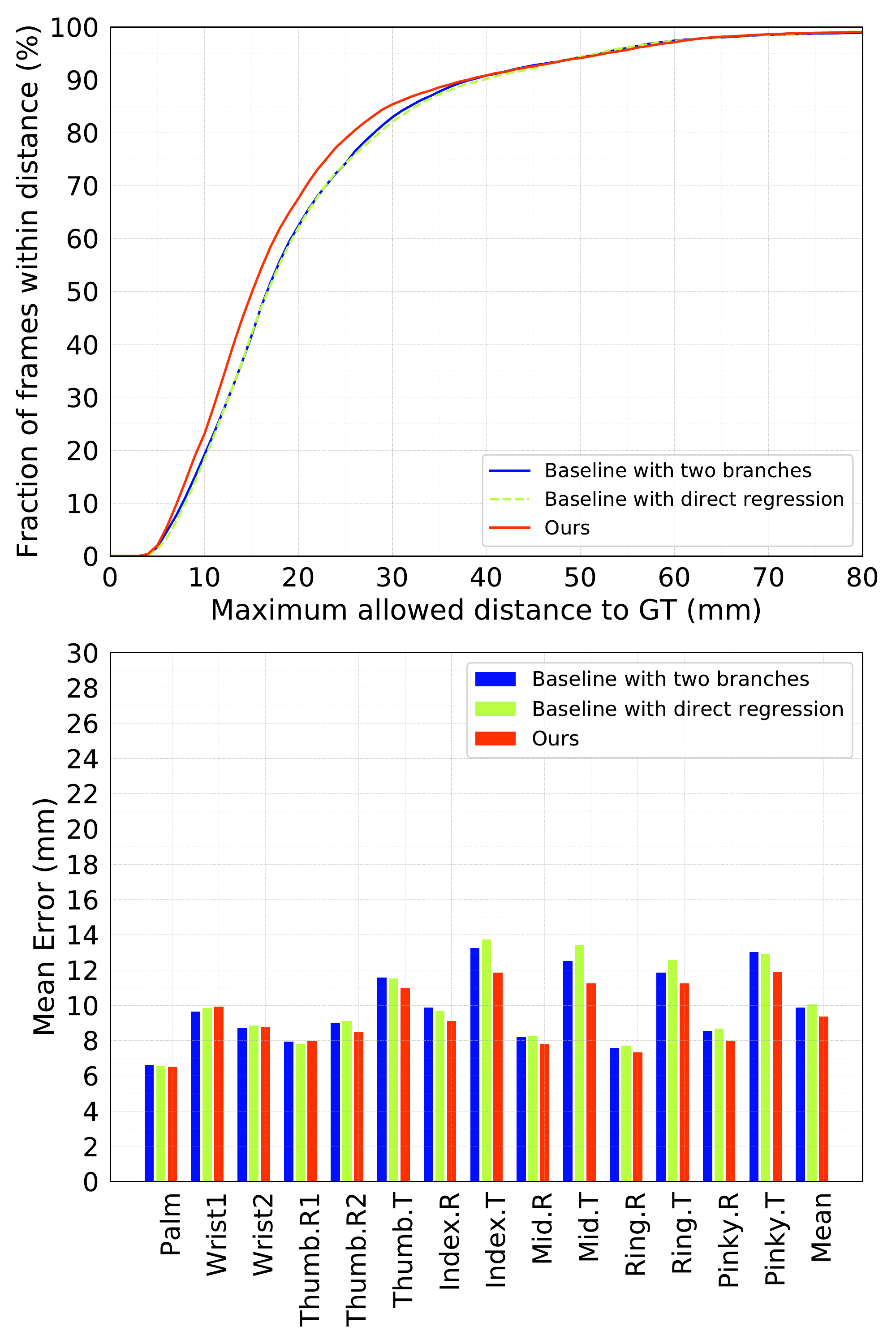}
{Self-comparison results on 3D distance errors (mm) per hand joint.\label{fig:self_result}}

\section{Experimental Results}
\subsection{Datasets and Evaluation Metrics}
We evaluated our network on the following three public hand pose datasets: ICVL~\cite{tang2014latent}, NYU~\cite{tompson2014real}, and MSRA~\cite{sun2015cascaded}.

\subsubsection{MSRA dataset}
The MSRA dataset~\cite{sun2015cascaded} contains 76k frames from nine different subjects with 17 different gestures. This dataset was captured with Intel’s Creative Interactive Gesture Camera~\cite{melax2013dynamics} and has 21 annotated joints, including 1 palm joint and four joints for each finger as shown in Fig.~\ref{fig:subset}(a). Following the most commonly used protocol~\cite{sun2015cascaded}, we used a leave-one-subject-out cross-validation strategy for evaluation on this dataset.

\subsubsection{NYU dataset}
The NYU dataset~\cite{tompson2014real} was captured with three Microsoft Kinects. It contains 72k training and 8k testing images from three different views. The training set was collected from one subject, while the testing set was collected from two subjects. According to the evaluation protocol that most previous works follow, we used only a frontal view and a subset of 14 annotated joints which is depicted in Fig.~\ref{fig:subset}(b) for both training and testing.

\subsubsection{ICVL dataset}
The ICVL dataset~\cite{tang2014latent} was captured with an Intel Realsense Camera. In this dataset, there are 22k frames from 10 different subjects for training and 1.5k images for testing. The training set includes an additional 300k augmented frames with in-plane rotations, but we did not use them because we applied online data augmentation during training, as described in Section~\ref{subsec:Implement}. 
This dataset has 16 annotated joints, including 1 palm joint and 3 joints for each finger, as shown in Fig.~\ref{fig:subset}(c).

\subsubsection{Evaluation metrics}
To evaluate the performance of the different 3D hand pose estimation methods, we used two metrics. The first metric is the average 3D distance error between the ground truth and predicted 3D position for each joint. The second one is the percentage of succeeded frames whose errors for all joints are within a threshold.

\begin{table}[t]
\caption{Comparison of the proposed method with state-of-the-art methods on three 3D hand pose datasets. Mean error indicates the average 3D distance error. And “$\ast$" means that ResNet-50 pre-trained on ImageNet is used as the backbone network.}
	\label{tab:porformance_comparison}
	\normalsize
	\begin{center}
	    \small
		\begin{tabular}{l c c c c}
			\hline
			\multirow{2}{*}{Method} & \multicolumn{3}{c}{Mean error (mm)} & \multirow{2}{*}{Input} \\
			\cline{2-4}
			 & ICVL & NYU & MSRA & \\
			\hline
			3D CNN~\cite{ge20173d} & -- & 14.1 & 9.58 & 3D\\
            SHPR-Net~\cite{chen2018shpr} & 7.22 & 10.78 & 7.96 & 3D\\
            3D DenseNet~\cite{ge2018real} & 6.7  & 10.6 & 7.9 & 3D\\
            Hand PointNet~\cite{ge2018hand} & 6.94 & 10.5 & 8.5 & 3D\\
            Point-to-Point~\cite{ge2018point} & 6.33 & 9.04 & 7.71 & 3D\\
            V2V-PoseNet~\cite{moon2018v2v} & 6.28 & 8.42 & 7.49 & 3D\\
            \hline
            Multi-view CNNs~\cite{ge2016robust} & -- & -- & 13.1 & 2D \\
		    DISCO~\cite{bouchacourt2016disco} & -- & 20.7 & -- & 2D\\
            DeepPrior~\cite{oberweger2015hands} & 10.4 & 19.73 & -- & 2D \\
            Feedback~\cite{oberweger2015training} & -- & 15.97 & -- & 2D \\
            Global2Local~\cite{madadi2017end} & -- & 15.60 & 12.8 & 2D\\
            CrossingNets~\cite{wan2017crossing} & 10.2 & 15.5 & 12.2 & 2D\\
            HBE~\cite{zhou2018hbe} & 8.62 & -- & -- & 2D\\
            REN (4x6x6)~\cite{guo2017region} & 7.63 & 13.39 & -- & 2D\\
            REN (9x6x6)~\cite{wang2018region} & 7.31 & 12.69 & 9.79 & 2D \\
            DeepPrior++~\cite{oberweger2017deepprior++} & 8.1 & 12.24 & 9.5 & 2D\\
            Pose-REN~\cite{chen2019pose} & 6.79 & 11.81 & 8.65 & 2D\\
            Generalized~\cite{oberweger2019generalized} & -- & 10.89 & -- & 2D\\
            CrossInfoNet~\cite{du2019crossinfonet} & 6.73 & 10.07 & 7.86 & 2D\\
            A2J*~\cite{xiong2019a2j} & 6.46 & 8.61 & -- & 2D \\
            \hline
            \bf{HCRNN (Ours)} & 6.54 & 9.37 & 7.70 & 2D\\
            \hline
		\end{tabular}
	\end{center}
\end{table}

\Figure[t!](topskip=0pt, botskip=0pt, midskip=0pt)[width=1.0\linewidth]{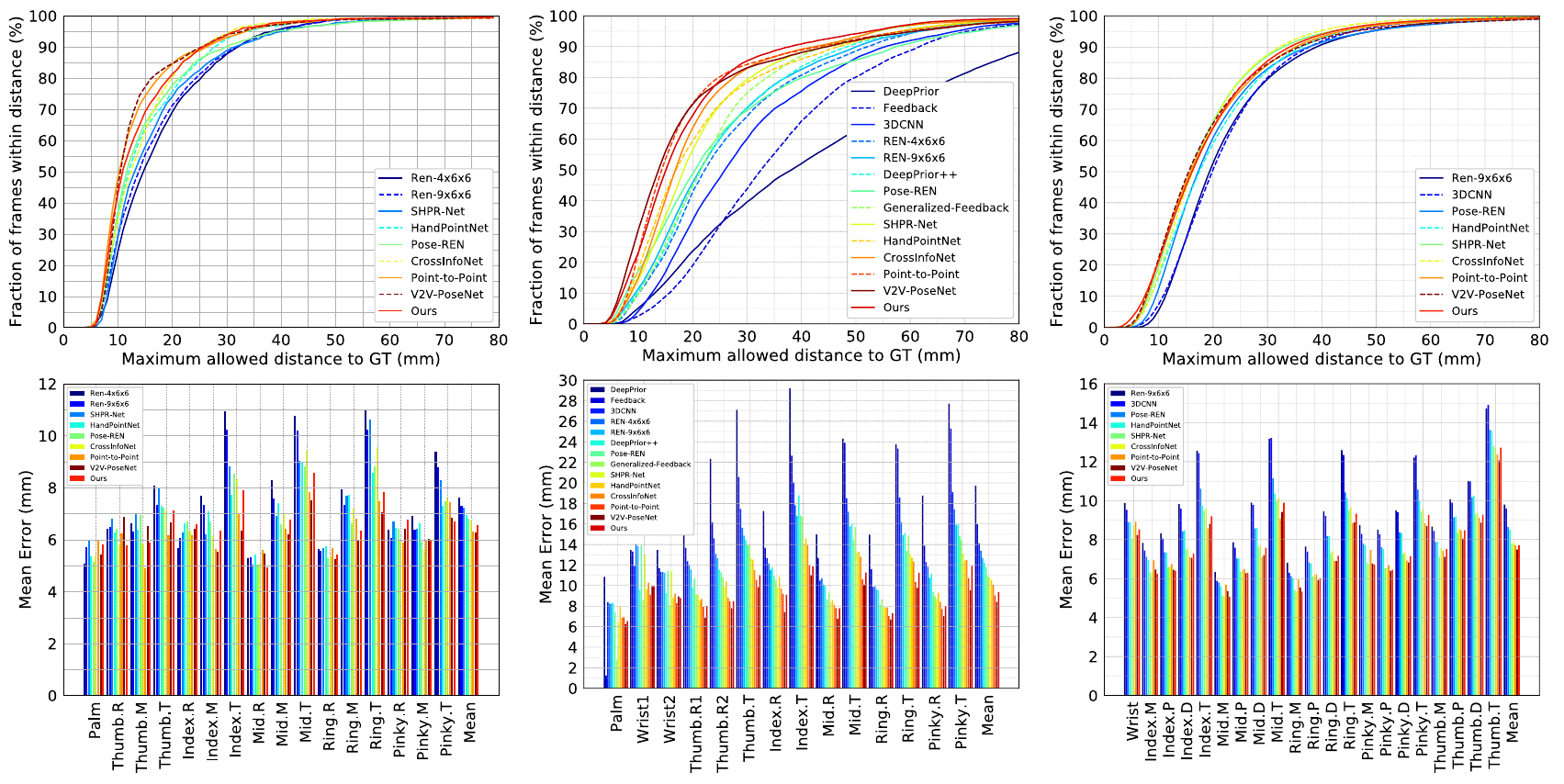}
{Comparison with state-of-the-art methods. Top row: percentage of success frames over different error thresholds. Bottom row: 3D distance errors per hand joint. Left: ICVL dataset, Center: NYU dataset, Right: MSRA dataset.\label{fig:integration}}

\subsection{Self-comparisons}
To analyze the contributions of each component of the proposed method, we conducted self-comparative experiments on the NYU~\cite{tompson2014real} dataset. First, we evaluated the effect of the number of branches. For this experiment, as shown in Fig.~\ref{fig:self}(a), we designed a two-branch network consisting of one branch for palm joint regression and the other one for unified finger joint regression, which is a similar approach in~\cite{du2019crossinfonet}. In other words, in the finger branch of the two-branch network, the RNN-based regression block estimates recurrently the joints of the five fingers simultaneously. For a fair comparison with the proposed architecture, we adjusted the number of convolution channels and dimensions of FC layers in the finger branch of the two-branch network so that each network has a similar number of parameters.
As shown in Fig.~\ref{fig:self_result}, the proposed network architecture with six branches performs better and reduces the mean 3D distance error~(mm) by 0.51~(from 9.88 to 9.37; see the last entry of the bar graph) as compared with the baseline network architecture. 
This result demonstrates that regressing all fingers jointly is inefficient because the five fingers are largely independent~\cite{sun2015cascaded}. By building a fine-grained branch for each finger, the network can learn richer features for finger pose estimation.

We also evaluated the effect of the RNN-based regression on finger joint estimation. We designed another network architecture, where the RNN-based regression block is replaced with FC layers, as shown in Fig.~\ref{fig:self}(b). In each finger branch of the FC-layer-based network, the output features are directly extracted from the input finger features with a simple FC layer instead of considering the structural properties of finger joints. As shown in Fig.~\ref{fig:self_result}, the proposed RNN-based network architecture achieves a better result than the FC-layer-based network with direct regression and reduces the mean 3D distance error~(mm) by 0.68 (from 10.05 to 9.37). These experiments confirm that the proposed RNN-based regression drives the network to utilize spatial dependencies between the sequential-joints of the finger for accurate 3d hand pose estimation. 

\Figure[t!](topskip=0pt, botskip=0pt, midskip=0pt)[width=0.99\linewidth]{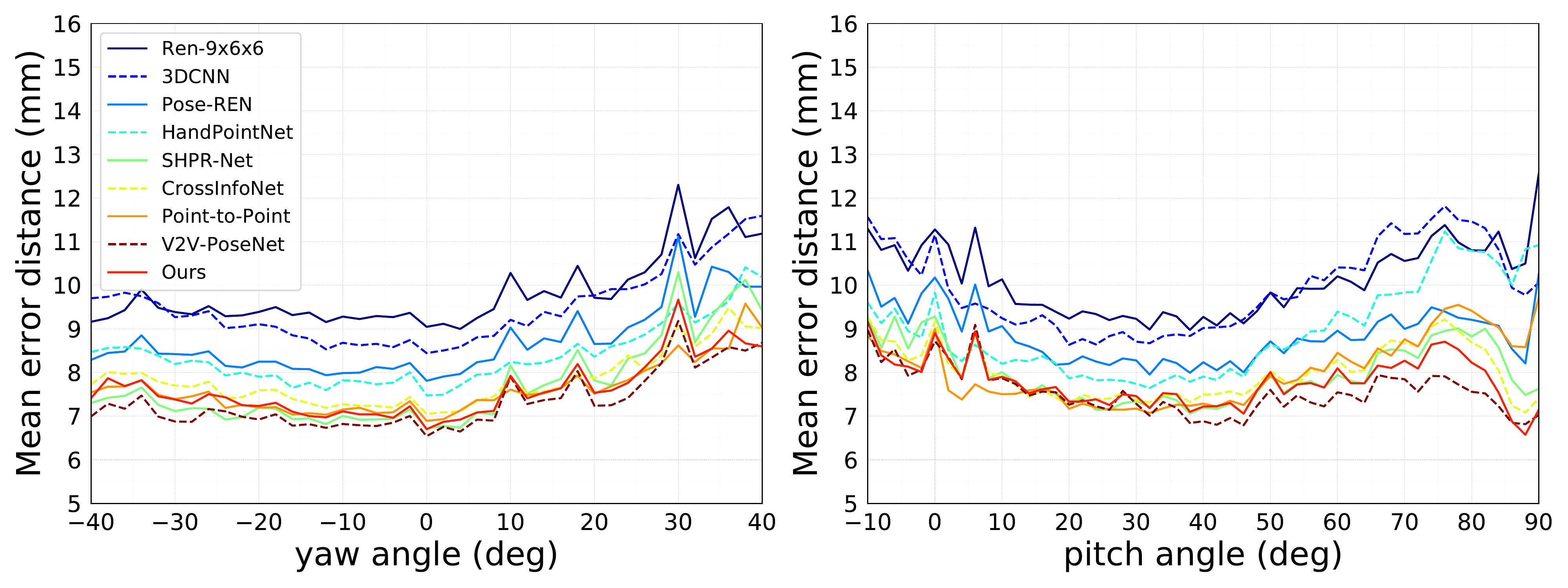}
{Comparison of mean error distance over different yaw (left) and pitch (right) viewpoint angles on MSRA dataset.\label{fig:msra_roll}}

\Figure[t!](topskip=0pt, botskip=0pt, midskip=0pt)[width=1.0\linewidth]{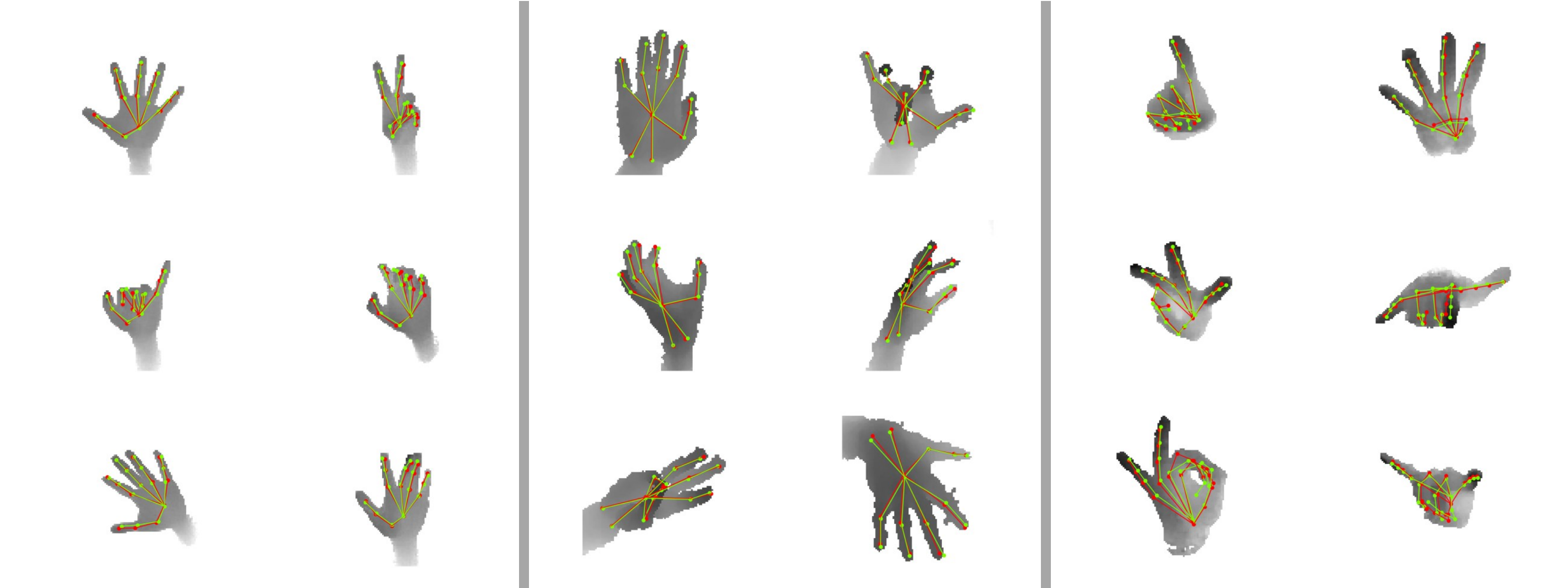}
{Qualitative results on the three public datasets. Left: ICVL dataset. Center: NYU dataset. Right: MSRA dataset. The ground truth is shown as red lines, and the prediction is shown as green lines.\label{fig:visual}}

\begin{table}[t]
\caption{Comparison of inference speed with state-of-the-art methods. The inference speed is measured with a single GPU.}
	\label{table:test time}
	\normalsize
	\begin{center}
		\begin{tabular}{l c c}
			\hline
			Method & Test speed (fps) & Input\\
			\hline
			V2V-PoseNet~\cite{moon2018v2v} & 3.5 & 3D\\
			Point-to-Point~\cite{ge2018point} & 41.8 & 3D\\
			HandPointNet~\cite{ge2018hand} & 48 & 3D\\
			3D DenseNet~\cite{ge2018real} & 126 & 3D\\
			3D CNN~\cite{ge20173d} & 215 & 3D\\
			\hline
			DeepPrior++~\cite{oberweger2017deepprior++} & 30 & 2D\\
			Generalized~\cite{oberweger2019generalized} & 40 & 2D\\
			CrossingNets~\cite{wan2017crossing} & 90.9 & 2D\\
			A2J~\cite{xiong2019a2j} & 105.1 & 2D\\
			CrossInfoNet~\cite{du2019crossinfonet} & 124.5 & 2D\\
			Feedback~\cite{oberweger2015training} & \bf{400} & 2D\\
			HCRNN (ours) & \underline{285} & 2D\\
			\hline
		\end{tabular}
	\end{center}
\end{table}

\subsection{Comparision with State-of-the-art Methods}
We compared the proposed network on three public 3D hand pose datasets with the most recently proposed methods using 2D depth maps as an input, including DISCO~\cite{bouchacourt2016disco}, DeepPrior~\cite{oberweger2015hands}, its improved version DeepPrior++~\cite{oberweger2017deepprior++}, Feedback~\cite{oberweger2015training},  Multi-view CNNs~\cite{ge2016robust}, REN-4x6x6~\cite{guo2017region}, REN-9x6x6~\cite{wang2018region}, Pose-REN~\cite{chen2019pose}, Generalized~\cite{oberweger2019generalized}, Global2Local~\cite{madadi2017end}, CrossingNets~\cite{wan2017crossing}, HBE~\cite{zhou2018hbe}, CrossInfoNet~\cite{du2019crossinfonet}, and A2J~\cite{xiong2019a2j}, as well as methods using 3D inputs, including 3D CNN~\cite{ge20173d}, SHPR-Net~\cite{chen2018shpr}, 3D DenseNet~\cite{ge2018real}, HandPointNet~\cite{ge2018hand}, Point-to-Point~\cite{ge2018point}, and V2V-PoseNet~\cite{moon2018v2v}. The average 3D distance error per joint and percentage of success frames over different error thresholds are respectively shown in Table~\ref{tab:porformance_comparison} and Fig.~\ref{fig:integration}. It is seen that our method outperforms most of the state-of-the-art methods with 2D inputs on all three datasets.
A2J shows the better accuracy on the NYU dataset; however, due to the adoption of a complex ResNet-50-based backbone network, it requires the higher computational complexity than the proposed method as described in Table~\ref{table:test time}.
As compared with the methods using 3D inputs, our method performs better than 3DCNN~\cite{ge20173d}, SHPR-Net~\cite{chen2018shpr}, HandPointNet~\cite{ge2018hand}, and 3D DenseNe~\cite{ge2018real} and achieves comparable performance with Point-to-Point~\cite{ge2018point} on the ICVL and MSRA datasets. On the NYU dataset, the results of the proposed method are worse than those of V2V-PoseNet~\cite{moon2018v2v} but are better in terms of percentage of success frame rates when the error threshold is larger than 30mm. On the MSRA dataset, following the evaluation protocol of prior works~\cite{chen2019pose, ge20173d, sun2015cascaded}, we also measured the mean joint error over various viewpoint angles. As shown in Fig.~\ref{fig:msra_roll}, our method can obtain promising results from large yaw and pitch angles, which demonstrates the robustness of our proposed method to viewpoint changes. The qualitative results of our method on three datasets are shown in Fig.~\ref{fig:visual}. It can be seen that our method can accurately estimate 3D hand joint locations on the three datasets.

\subsection{Runtime}    
Because 3D hand pose estimation tasks play an important role as a sub-system in the overall human-computer interaction (HCI) system such as in-vehicle infotainment, the inference speed is an important factor for the practical application. Table~\ref{table:test time} compares the inference speed of conventional and the proposed methods on a single GPU. While top-ranked methods using 3D inputs have a higher inference time owing to the time-consuming 3D convolution operation or data conversion procedure, our method has a faster inference speed owing to its efficiency of 2D CNN-based architecture. The proposed HCRNN is ranked in 2nd place among the compared methods behind Feedback. Based on the aforementioned results, it can be seen that our proposed HCRNN not only achieves competitive performance compared with state-of-the-art methods but also is very efficient, having a high frame rate, which shows the applicability to real-time applications. 

\section{Conclusion}
To design a practical architecture for 3D hand pose estimation, we considered the articulated structure of the hand and proposed an efficient regression network, namely termed HCRNN. The proposed HCRNN has a hierarchical architecture, where six separate branches are trained to estimate the position of each local part of the hand: the palm and five fingers. In each finger branch, we adopted an RNN to model spatial dependencies between the sequential-joints of the finger. In addition, HCRNN is built on a 2D CNN that directly takes 2D depth images as inputs, making it more efficient than 3D CNN-based methods. The experimental results showed that the proposed HCRNN not only achieves competitive performance compared with state-of-the-art methods but also has a highly efficient running speed of 285 fps on a single GPU.

\bibliographystyle{IEEEtran}
\bibliography{refs}

\newpage
\begin{IEEEbiography}[{\includegraphics[width=1in,height=1.25in,clip,keepaspectratio]{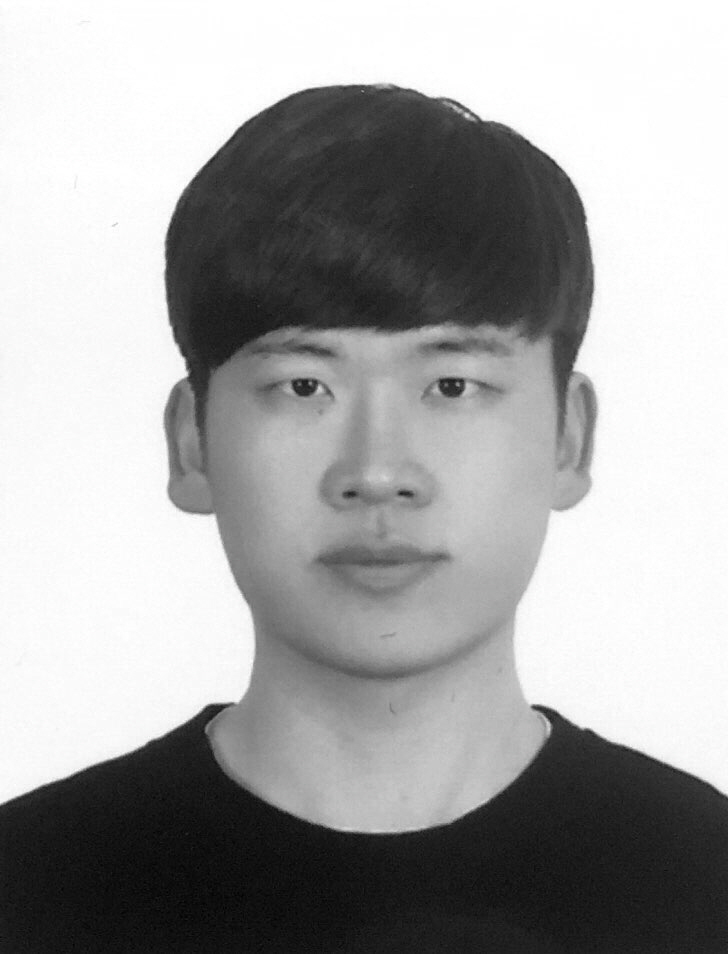}}]{Cheol-Hwan~Yoo}
received the B.S. degree in electrical engineering from Korea University in 2014, where he is currently pursuing the Ph.D.degree in electrical engineering. His research interests include deep learning, image processing, and computer vision.
\end{IEEEbiography}

\begin{IEEEbiography}[{\includegraphics[width=1in,height=1.25in,clip,keepaspectratio]{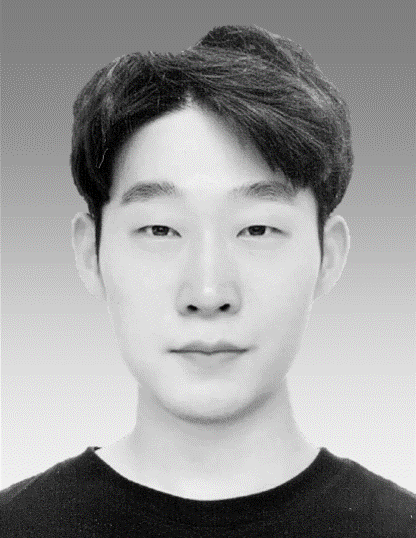}}]{Seo-Won~Ji}
received the B.S. degree in electrical engineering from Korea University, Seoul, South Korea, in 2015, where he is currently
pursuing the Ph.D. degree in electrical engineering. His research interests are the areas of image processing, computer vision, and deep-learning.
\end{IEEEbiography}

\begin{IEEEbiography}[{\includegraphics[width=1in,height=1.25in,clip,keepaspectratio]{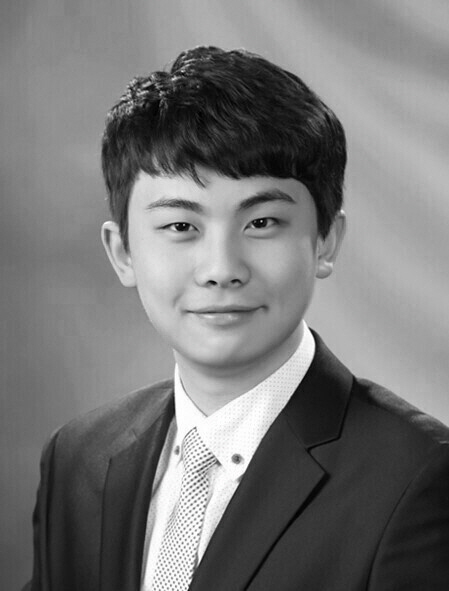}}]{Yong-Goo~Shin}
received the B.S. and Ph.D. degrees in Electronics Engineering from Korea University, Seoul, Rep. of Korea, in 2014 and 2020, respectively. He is currently a research professor in the Department of Electrical Engineering of Korea University. His research interests are in the areas of digital image processing, computer vision, and artificial intelligence.
\end{IEEEbiography}

\begin{IEEEbiography}[{\includegraphics[width=1in,height=1.25in,clip,keepaspectratio]{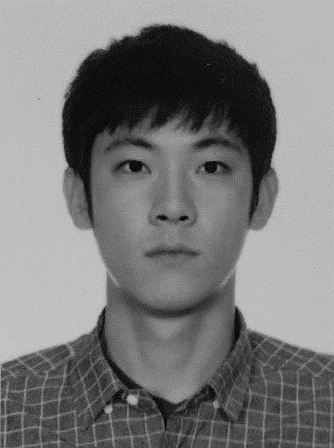}}]{Seung-Wook~Kim}
received the B.S. and Ph.D. degrees in electronics engineering from Korea University in 2012 and 2019, respectively. He is currently a research professor in the School of Electrical Engineering, Korea University, Seoul, Korea. His research interests are deep-learning-based applications including image processing and computer vision.
\end{IEEEbiography}

\vfill

\begin{IEEEbiography}[{\includegraphics[width=1in,height=1.25in,clip,keepaspectratio]{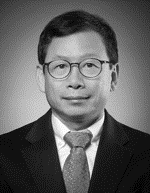}}]{Sung-Jea~Ko}
(M’88-SM’97-F’12) received his Ph.D. degree in 1988 and his M.S. degree in 1986, both in Electrical and Computer Engineering, from State University of New York at Buffalo, and his B.S. degree in Electronic Engineering at Korea University in 1980. In 1992, he joined the Department of Electronic Engineering at Korea University where he is currently a Professor. 

From 1988 to 1992, he was an Assistant Professor in the Department of Electrical and Computer Engineering at the University of Michigan-Dearborn. He has published over 210 international journal articles. He also holds over 60 registered patents in fields such as video signal processing and computer vision.

Prof. Ko is the 1999 Recipient of the LG Research Award. He received the Hae-Dong best paper award from the Institute of Electronics and Information Engineers (IEIE) (1997), the best paper award from the IEEE Asia Pacific Conference on Circuits and Systems (1996), a research excellence award from Korea University (2004), and a technical achievement award from the IEEE Consumer Electronics (CE) Society (2012). He received a 15-year service award from the TPC of ICCE in 2014 and the Chester Sall award (First Place Transaction Paper Award) from the IEEE CE Society in 2017. He has served as the General Chairman of ITC-CSCC 2012 and the General Chairman of IEICE 2013. He was the President of the IEIE in 2013, the Vice-President of the IEEE CE Society from 2013 to 2016, and the distinguished lecturer of the IEEE from 2015 to 2017. He is a member of the National Academy of Engineering of Korea. He is a member of the editorial board of the IEEE Transactions on Consumer Electronics. 
\end{IEEEbiography}

\EOD

\end{document}